\def\year{2022}\relax

\documentclass[letterpaper]{article} %
\usepackage[inline]{enumitem}

\usepackage{todonotes}

\usepackage{adjustbox}

\usepackage[utf8]{inputenc}
\usepackage{pgfplots}
\DeclareUnicodeCharacter{2212}{−}
\usepgfplotslibrary{groupplots,dateplot}
\usetikzlibrary{patterns,shapes.arrows}
\pgfplotsset{compat=newest}
\usepackage{aaai22}  %
\usepackage{times}  %
\usepackage{helvet}  %
\usepackage{courier}  %
\usepackage[hyphens]{url}  %
\usepackage{graphicx} %
\usepackage{quotes}
\usepackage{multirow}
\usepackage{booktabs}
\usepackage{array}
\usepackage{arydshln}
\usepackage{caption}
\usepackage{subcaption}
\usepackage{amsmath}
\usepackage{cleveref}
\usepackage[hang,flushmargin]{footmisc} %

\usepackage{natbib}  %
\usepackage{caption} %
\DeclareCaptionStyle{ruled}{labelfont=normalfont,labelsep=colon,strut=off} %
\usepackage{algorithm}
\usepackage{algorithmic}

\newcommand{\pseudosection}[1]{\textit{{#1.}}}
\newcommand{\pseudosectionp}[1]{\textit{{#1}}}
\newcommand{\gld}{GoLD}

\usepackage{fancyhdr}

\usepackage{newfloat}
\usepackage{listings}
\floatstyle{ruled}
\newfloat{listing}{tb}{lst}{}
\floatname{listing}{Listing}
\title{Bridging the Gap: Using Deep Acoustic Representations to Learn Grounded Language from Percepts and Raw Speech}
\author{Gaoussou Youssouf Kebe,\textsuperscript{\rm 1} Luke E. Richards,\textsuperscript{\rm 1,2} Edward Raff,\textsuperscript{\rm 1,2} Francis Ferraro,\textsuperscript{\rm 1} Cynthia Matuszek\textsuperscript{\rm 1}}
\affiliations {
    \textsuperscript{\rm 1}Department of Computer Science and Electrical Engineering (CSEE)\\
    University of Maryland, Baltimore County\\
    1000 Hilltop Circle\\
    Baltimore, MD 21250\\
    \textsuperscript{\rm 2} Booz Allen Hamilton\\
    National Business Park, 304 Sentinel Drive\\
    Annapolis Junction, MD 20701\\
    gaoussou1@umbc.edu, lerichards@umbc.edu, raff\_edward@bah.com, ferraro@umbc.edu, cmat@umbc.edu
}
\author{Paper ID: 2679}
\author{Gaoussou Youssouf Kebe,\textsuperscript{\rm 1} Luke E. Richards,\textsuperscript{\rm 1,2} Edward Raff,\textsuperscript{\rm 1,2} Francis Ferraro,\textsuperscript{\rm 1} Cynthia Matuszek\textsuperscript{\rm 1}}

\begin{document}

\maketitle
\thispagestyle{fancy}   %

\begin{abstract}
Learning to understand grounded language, which connects natural language to percepts, is a critical research area. Prior work in grounded language acquisition has focused primarily on textual inputs. In this work we demonstrate the feasibility of performing grounded language acquisition on paired visual percepts and raw speech inputs. This will allow interactions in which language about novel tasks and environments is learned from end users, reducing dependence on textual inputs and potentially mitigating the effects of demographic bias found in widely available speech recognition systems. We leverage recent work in self-supervised speech representation models and show that learned representations of speech can make language grounding systems more inclusive towards specific groups while maintaining or even increasing general performance.
\end{abstract}
\section{Introduction}

Learning to understand grounded language---learning the semantics of language that occurs in the context of, and refers to, the broader world---is a rich area of work that has engaged researchers from robotics~\cite{tellex2020robotslanguage}, natural language processing~\cite{liu2016jointly}, vision~\cite{deng2018cvpr}, and cognitive science~\cite{salvucci2021interactive}, among others. In robotics, grounded language refers primarily to grounding human utterances in the perceived physical world of objects, actions, and the environment. Learning from grounded language is an intuitive choice for interacting with agents in a physical environment.

While language learning offers a clearly defined way for embodied agents to learn about changing environments and goals directly from a specific end user, with some exceptions, the majority of current work in this area still operates primarily on textual data. This approach significantly limits our ability to deploy agents in realistic human environments, where spoken inputs can be expected. Existing work on using speech directly typically relies on off-the-shelf speech-to-text systems. These systems are rarely developed in tandem with the robotics community, and so do not take the unique challenges of robotic sensing into account~\cite{marge2021spoken}. In addition, current ASR systems work inconsistently across demographics~\cite{tatman2017gender,hinsvark2021accented}, which represents a problem in inclusive design. They are also ``black box'' systems that cannot improve their speech recognition from their environment or other perceptual clues. Since the grounding system does not have access to the information used by these models, it can only rely on their sometimes erroneous output.

In this work, we bridge the gap between learning grounded language about the perceived world via text-based language, and directly learning to recognize speech without access to the physical context in which it occurs. We contribute a detailed analysis of natural language grounding from raw speech to robotic sensor data of everyday objects using state-of-the-art speech representation models. We then conduct an analysis of audio and speech qualities of individual participants, in which we demonstrate that learning directly from raw speech mitigates the performance difference between linguistic groups on a well-known grounded language learning problem.

The primary contributions of this paper are twofold. \textbf{First,} we demonstrate the feasibility of acquiring grounded language directly from end-user speech using a relatively small number of data points, without relying on intermediate textual representations. \textbf{Second,} we show that such learning improves performance on users with accented speech as compared to relying on automatic transcriptions.

The remainder of this paper is organized as follows. After a discussion of related work, we describe our approach, including the dataset~\cite{kebe2021neurips}, learning method, and different speech-based features tested, as well as the object selection task we use to determine whether language groundings have been learned successfully. We compare the experimental results of learning from percepts and raw speech directly, vs. the traditional transcription-first approach, and provide an analysis of both approaches when learning from spoken language from different demographics present in the dataset.

\section{Related Work} \label{sec:related}

\pseudosection{Grounded Language Acquisition from Text} %
In robotics, language is grounded in real-world actions and percepts, whose applications include following task instructions~\cite{vanzo2020grounded, bastianelli2016discriminative,shridhar2020ingress}, navigation instruction following,~\cite{Shah2018FollowNetRN,zang-etal-2018-translating}%
, and learning groundings from human-robot dialog~\cite{thu2017symbol,thomason:jair20}%
, among others. While some of these approaches use text derived from ASR, none use speech directly.
The focus of this paper is on learning language groundings directly from speech; we demonstrate this work on the common grounding problem of object retrieval~\cite{nguyen2020robot,hu2016natural}.%

In vision, grounded language typically refers to how language refers to existing images. From image and video captioning~\cite{kinghorn2019hierarchical, wang2018reconstruction,chen2019unsupervised} to large-scale pre-training~\cite{lu2019vilbert}%
, learning from vision-language pairs is an active field of research.
In this work, we use the manifold alignment approach of \citet{Nguyen2021Practical}, in which language and vision representations are projected into a shared manifold, which is used to retrieve relevant objects given a natural language description. The novelty of our work is not in the triplet loss learning method for multi-modal alignment but in the comparison of transcription-based versus raw speech methods, and analysis of performance for end-users.

\pseudosection{Spoken grounded language learning} %
While the majority of existing grounded language learning is performed over text sources (either typed or transcribed), there are exceptions that demonstrate the importance of learning directly from speech. In early work, \citet{roy2003journal} presented a grounded speech learner that segments words from continuous speech. Our problem is more complex, in that we aim to ground full descriptions rather than words. The work most closely related to our research explores using audio-visual neural networks to learn semantic similarity between single images and raw spoken utterances~\cite{Harwath_2018_ECCV}. By contrast, we focus on multi-frame RGB-D percepts gathered from a sensor, aiming  to identify individual objects rather than entire images.
Our work is most similar to that of Chrupa{\l}a et al.~\cite{chrupala2018symbolic,Chrupa_a_2017} and \citet{zhang2020sound}. However, the speech corpora in those works are collected by asking speakers to read captions of images aloud. This may remove grammatical constructs, disfluencies, and speech repair, effectively gating the complexities of speech through written language. The dataset used in our work consists purely of people describing objects.

We are not aware of previous work comparing grounding from raw speech to the widely used transcription-first approach. Additionally, we show how to create a speech-based grounding system based on complex perceptual data using a comparatively small number of data points, which is consistent with the requirements and available resources for implementing on robotic systems. Compared to previous work, we leverage depth information and pretrained speech representation models to ground naturalistic spoken language in a model which converges with fewer data pairs; \citet{Harwath_2018_ECCV} used 402,385 image-caption pairs and \citet{Chrupa_a_2017} specifically mention that the Flickr8K dataset of 40,000 image-caption pairs is small for the speech task.
Finally, we avoid the computational overhead of fine-tuning the model for specific domains, which may change as the robot experiences new environments.

\pseudosection{Language and Speech in Robotics} %
The role of language in robotics is wide-ranging~\cite{tellex2020robotslanguage}, and the role of speech in particular is starting to receive significant attention~\cite{marge2021spoken}.
Natural language is widely used in HRI tasks, for example in dialogue with assistive robots~\cite{kulyukin2006natural} or to facilitate human learning~\cite{lee2011effectiveness,pazylbekov2019similarity, scassellati2018teaching, kose2015effect, Ramachandran_thinking}.
Speech-based HRI, in particular, has been applied to a wide variety of problems, such as emotion recognition~\cite{fischer2019emotion,williams2019aida}, social robotics~\cite{mollaret2016multi,al2014human},
and speech recognition. \Citet{mead2016autonomous} determine how a robot should position itself for optimal speech and gesture recognition, complementing work on how people expect a robot to react when given instructions~\cite{moolchandani}. Speech is also an important source of insight into how different groups interact with robots, for example in assessing the communications of dementia patients through speech features such as pitch~\cite{yamanaka2016assessing}.

\pseudosection{Speech Processing Bias} %
While much previous work relies on ASR systems, these systems have known biases in their ability to recognize speech without errors. For now, most widely available ASR approaches depend on large-scale data~\cite{zhang2009unsupervised}. These large datasets are usually derived from fairly heterogeneous groups~\cite{koenecke2020racial}. Given this, we see
gender~\cite{alsharhan2020investigating,tatman2017gender},
race~\cite{blodgett2017racial}, disability status~\cite{fok2018towards}, and native language/dialect~\cite{hinsvark2021accented} disparities in successful ASR, reducing technology accessibility for under-represented groups. Contemporaneous work \cite{liu2021towards} introduced a dataset to measure ASR performance across age, gender and skin type.  Technical remediation approaches remain scarce~\cite{meyer2020artie,tan2020s}, although there has been some work on multilingual grounded language~\cite{kery2019ROMAN}.
To our knowledge, no previous work in grounded language acquisition has examined the impact of these factors in speech.

\section{Data and Embedding Representations}\label{sec:data}

In this work, we discuss grounding language using vision and depth percepts paired with either ASR-transcribed language, or with raw speech inputs. In this section, we describe the dataset and speech featurizations we consider.

\subsection{Dataset}\label{dataset}

We use the \gld\ dataset~\cite{kebe2021neurips}, which includes 207 unique objects from 47 object classes. Each object in the dataset has on average 4 instances of RGB-D (image plus depth) data from different angles. Color aligned with depth images offer greater information on the objects properties that are critical for downstream manipulation tasks commonly performed within robotics. Due to the lower resource environment robots are deployed in, performing sensor fusion enables rich multimodal object representations with less data. The \gld\ dataset contains a total of 16,500 raw sound files (with transcriptions) from 552 distinct speakers. We manually annotated speakers with seven different traits describing speaker and sound file characteristics (see \cref{ssec:useranalysis}).

\subsection{Perceptual Representations}\label{subsec:features}

\begin{figure*}[t]
\centering
\resizebox{.8\textwidth}{!}{
\includegraphics{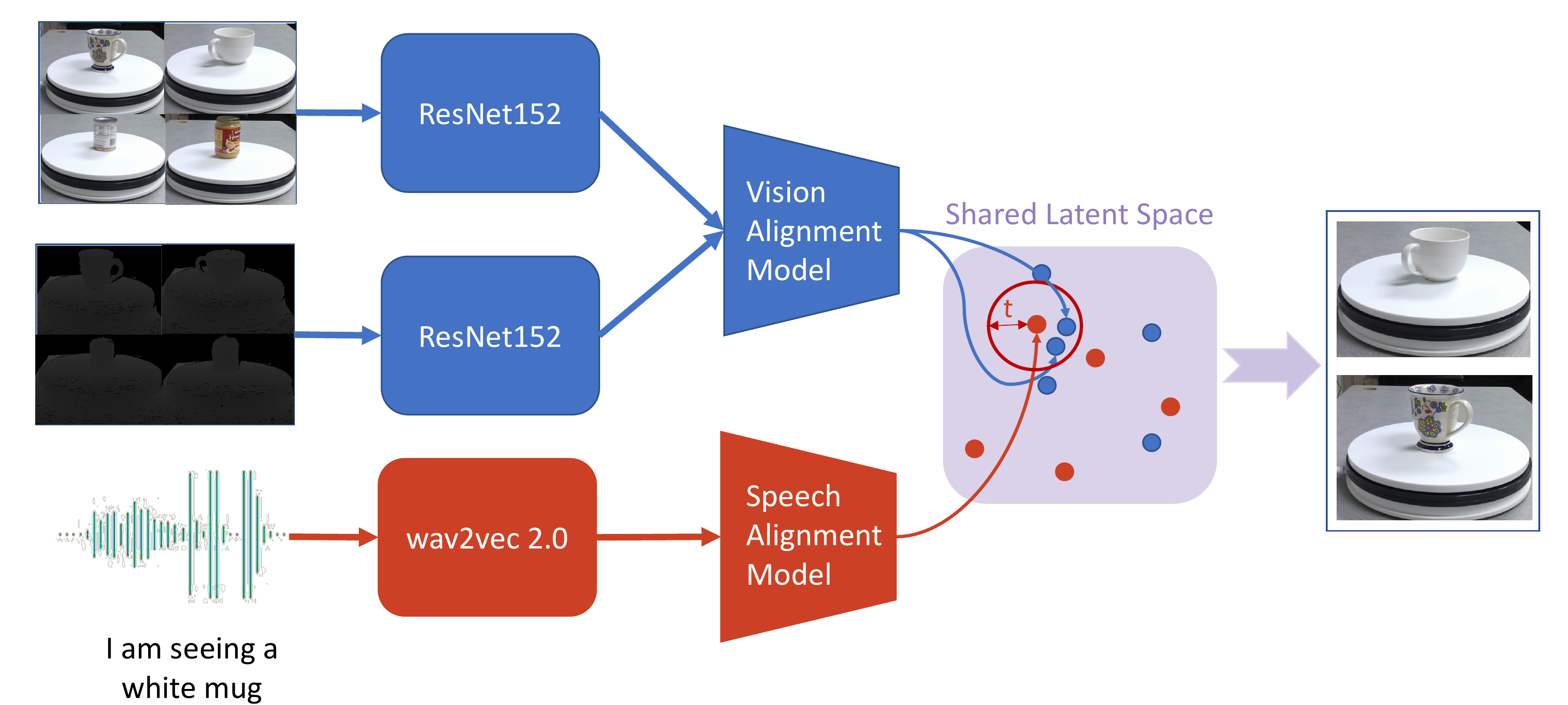}
}
\caption{
Our learning approach is to use manifold alignment in an attempt to capture a manifold between speech and visual perception. This approach is applied to grounded language acquisition by projecting visual and language representations into a shared latent space, where projections from both domains are closer to other projections of the same class. For example, the projection of the language utterance ``I am seeing a white mug'' should be close to the projection of the visual percepts of a white mug.
}
\label{fig:architecture}
\end{figure*}

In order to learn language from perceptual inputs, all modalities (RGB-D, transcribed text, and speech) must be featurized appropriately. In particular, because handling raw speech as perceptual input for grounding is a novel task, we experiment with multiple speech representation models. To avoid overfitting on this relatively small dataset, we did not fine-tune hyperparameters for feature extractions.

\subsubsection{Visual Representations}
Visual features are extracted using ResNet152 pre-trained on ImageNet~\cite{he2016deep}, which achieves very strong results in image classification and object detection tasks (as a result, our system depends indirectly on labeled data by way of this pre-training). The last fully connected layer is removed to obtain the 2048-dimensional features used for classification. Both RGB and depth are processed through this network, the latter by colorizing depth images~\cite{richards2020iros}. This yields two 2048-dimensional vectors, which are concatenated to create a multimodal object representation.

\subsubsection{Transcribed Text Representation}
\label{sssec:transcriptions}
Language features for transcribed text are obtained using BERT, a self-supervised bidirectional language model that achieves state-of-the-art performance in multiple NLP tasks~\cite{bert}. BERT's embeddings and linguistic performance make it useful for clustering~\cite{jawahar2019acl},
making BERT more appropriate for sentence-based language grounding than the commonly used words-as-classifiers approaches~\cite{schlangen2016resolving}. For a given natural language description, we obtain a 3,072 dimensional vector by extracting the average representation across the last four layers. We consider transcriptions obtained from wav2vec 2.0~\cite{wav2vec2} as it has been shown to achieve near state-of-the-art performance~\cite{librispeech}. These transcriptions are also directly comparable with our speech-based methods.

\subsubsection{Speech Representation}
We consider three different self-supervised speech models, which have recently shown success in phoneme and speech recognition~\cite{decoar,vq-wav2vec,wav2vec2}. The speech representations extracted from these models are intended to encode semantic information directly captured from raw speech; this is precisely the informational core that language acquisition seeks to capture. We argue that the process of mapping raw speech representations to discrete transcriptions leads to a loss of information that may be detrimental to the performance of the grounding model. Therefore, we expect learning directly from the representations extracted from these models to reduce the effect of speech recognition errors on human-robot communication.
We consider a state-of-the-art model speech model, wav2vec 2.0~\cite{wav2vec2}, and two other near state-of-the-art models in vq-wav2vec~\cite{vq-wav2vec} and DeCoAR~\cite{decoar}. We expect the performance of these three models will provide insights into the progress made and the overall direction of the field of acoustic representation learning.

\pseudosectionp{Baseline: Mel-frequency cepstral coefficients}
(MFCCs)~\cite{mfcc} are a na\"ive baseline, which are widely used and easy to implement. MFCCs are inspired by the human auditory system and are extracted via a Discrete Fourier Transform analysis. They are frequently used in speech recognition systems, providing an effective comparison to using such systems directly. We use this baseline to evaluate how the highly pre-trained speech representation models compare to a simple speech feature extractor.

\pseudosectionp{Model 1: DeCoAR}~\cite{decoar} is inspired by the vector-based word representation ELMo~\cite{elmo}. In ELMo, word vectors are learned from a contextualized bidirectional language model that is pre-trained on a large text corpus to predict the next word given a context. Unlike unidirectional language models, ELMo considers context from both directions. DeCoAR is an LSTM-based model that takes inspiration from ELMo's bidirectionality to learn deep contextualized acoustic representations, applying the same idea to speech by predicting a given slice of sound using past and future context through a backward and a forward LSTM. The sound is represented as sequential filterbank features. %
Combining these features, DeCoAR attempts to predict a given slice of sound by considering context from $K$ steps ahead and behind. We use existing pre-trained weights for DeCoAR.\footnote{github.com/awslabs/speech-representations}%

\pseudosectionp{Model 2: vq-wav2vec}~\cite{vq-wav2vec} is based on wav2vec, a word2vec~\cite{mikolov2013distributed} inspired convolutional network, pre-trained on a context-prediction task to learn representations of audio data. The model outputs a series of 512 dimensional vectors representing 30ms of sound data with a stride of 10ms. The vq-wav2vec approach uses a two-step process: first, wav2vec is remodeled through quantization to output discrete units of speech, which are then fed through a BERT model pre-trained on speech signals to output final speech representations.
This process results in two pre-trained models (vq-wav2vec and BERT). We use the authors' pre-trained weights for both models.\footnotemark

\pseudosectionp{Model 3: wav2vec 2.0}~\cite{wav2vec2} builds upon vq-wav2vec~\cite{vq-wav2vec} in two key ways. First, the output of a wav2vec-like feature encoder is fed into a transformer. Second, quantization is used to discretize the feature extractor's output. However, the continuous representation produced by the feature extractor is fed into the transformer and the quantized representation is only used as target output during pre-training. Accordingly, the transformer learns over continuous representations rather than the discrete ones used in vq-wav2vec. The objective function is similar to BERT's~\cite{bert} masked language modeling. Similarly to vq-wav2vec, we use pre-trained weights.\footnotemark[\value{footnote}] The pre-trained model was also fine-tuned for speech recognition using Connectionist Temporal Classification (CTC)~\cite{CTC}. The same process is used to extract the transcriptions mentioned earlier in the transcribed text representation section.\footnotetext{github.com/pytorch/fairseq/tree/master/examples/wav2vec}

\section{Approach}\label{sec:approach}
We approach the problem of learning groundings from unconstrained speech in an unconstrained environment. Our primary visual percepts are RGB-D point clouds obtained from a mounted Kinect 3. We encode RGB and depth using a sensor fusion convolution neural network for both RGB and depth~\cite{eitel2015multimodal,richards2020iros} and experiment with learning from various pre-trained speech representation models, as well as from transcriptions. Our learning approach is to use manifold alignment with triplet loss~\cite{Nguyen2021Practical} in an attempt to capture a manifold between speech and visual perception. This manifold represents the grounding between query language and objects in a selection task.%

\subsection{Learning}  \label{ssec:learning}
We use the manifold alignment approach of \citet{Nguyen2021Practical}. Given two heterogeneous representations, the goal is to learn mappings to a shared latent space. This approach is applied to grounded language acquisition by projecting visual and language representations into a shared high-dimensional latent space, in which the projection of a language utterance that describes an object $o$ of class $c$ should be `close' to the projections of other language utterances and visual percepts belonging to $o$, and to a broader degree other objects of class $c$, as seen in Fig.~\ref{fig:architecture}.

\pseudosectionp{Triplet Loss} \label{ssec:triplet-learning}
is a popular geometric approach that has shown success in learning metric embeddings~\cite{facenet,HermansBL17,Dong_2018_ECCV}. Learning uses triplets of the form ${(a, p, n)}$, where $a$ is an `anchor' point, $p$ is a positive instance of the same class as the anchor (e.g., mug), and $n$ is a negative instance from a different class (e.g., apple). For each triplet, the embedding function $f$ is learned so that the distance between $a$ and $n$ is maximized while the distance between $a$ and $p$ is minimized. This is achieved via the following loss function:
\begin{equation}
\begin{split}
L = \max(d(f(a, m_a), f(p, m_p)) - d(f(a, m_a),\\ f(n, m_n)) + \alpha, 0)
\end{split}
\end{equation}
where $d$ is a distance metric, $m_x$ is the modality of point $x$, and $\alpha$ is a margin imposed between positive and negative instances. This approach lends itself well to a human learning scenario, in which a person could provide positive and negative examples of a given description or object.

Due to the heterogeneous nature of our problem, the embedding function $f$ is the encoder that projects instances of a given modality into the shared manifold. We implement a different encoder for each modality as the input size and type are different. Each member of the triplet ${(a, p, n)}$ can be selected from the vision or language domain. The domain is randomly selected. We use cosine distance as the distance metric and a margin $\alpha = 0.4$.

\pseudosection{Training}
We split our 16,500 pairs of RGB-D data and descriptions into training, validation and testing sets of respectively 13,040; 1,620; and 1,840 instances.
We train alignment models with five different language representations: BERT embeddings for wav2vec 2.0 transcriptions in addition to MFCCs, vq-wav2vec, wav2vec 2.0 and DeCoAR embeddings for raw speech. All five are aligned with the visual features. All pre-trained feature extraction models are fixed during training. Only alignment models are optimized.
The default architecture of our alignment model is comprised of language and vision sub-networks that both consist of an input layer, two hidden layers with rectified linear units (ReLu) as activation functions and an output layer to obtain a final 1024 dimensional projection into the shared manifold. We use this architecture for the BERT, wav2vec 2.0, vq-wav2vec and DeCoAR embeddings.

Because of the low-dimensional nature of MFCCs, we also consider an LSTM-based language network. Instead of mean pooling, we input sequential MFCCs into a LSTM with 64 dimensional hidden states and concatenate the last 32 hidden states together resulting in a 2048 dimensional vector, which is input to a fully connected layer to obtain the final projection.
All five methods are trained for a total of 300 epochs using Adam \cite{kingma2014adam} with a learning rate of 0.001 that we reduce by a factor of 10 after each hundred epochs.

\subsection{User Trait-based Analysis}
\label{ssec:useranalysis}

One of the goals of speech-based and other machine learning technologies is that they should be accessible, fair, and unbiased towards various demographics of users. In the following section, we outline how we analyze differences in outcomes between transcription-based text versus raw speech approaches for a variety of speaker traits.

\subsubsection{ Individual User Analysis}
\label{sssec:singleuser}

For embodied learning systems to be deployed effectively, they must be able to learn from individual users with a variety of speaker characteristics. To analyze the ability of the system to learn from individuals with a variety of speaker traits, we label speakers in the \gld\ dataset based on qualities in which speaker variance is known to affect the success of speech recognition models, \textit{e.g.}, accented \textit{vs.} unaccented speech. We then compare the results obtained when using wav2vec 2.0 speech representations and wav2vec 2.0 transcriptions.

In the dataset, we are able to analyze individual users in the context of providing spoken learning examples. We define a user as a unique participant in the Amazon Mechanical Turk (AMT) task. For testing the effectiveness of learning from individuals, we restrict this first evaluation to users who provide sufficient exemplars (described in \ref{sssec:singleuserlearning}). Each user contributed a variety of examples, each with idiosyncrasies and unique perceptions of the description task. This offers a diverse set of user interactions.
We examine the qualities of the vocal samples by randomly sampling five speech events per user and annotating speakers based on perceived gender (man, woman, or undetermined),\footnote{Gender and sex are complex constructs. We asked annotators to choose the category that seemed to `best describe' the speaker, but acknowledge the limitations of this approach.} the presence of a non-American-English accent, creak (a raspy vocal sound), hoarseness (a strained vocal sound),  muffled-ness (obstruction in vocal event), volume (a range from 1--4, with 2 being average volume), and level of background noise (1--4, with 1 being average and 4 being high). Some of these traits may vary for the same user from one example to the next. With that in mind, the annotations for those traits were done to reflect the majority case and may not apply to all examples provided by the user.
We then evaluate by splitting each individual user's data into training and test splits and testing the learning system's ability to learn successfully from individual speakers.

\subsubsection{User-Group Analysis}

To better understand the extent that user-specific traits affect the performance of the learned model, in our second evaluation, we train over groups of multiple users with shared characteristics.
In this analysis, we split based on perceived gender, accent, muffled-ness, background noise and volume.
The accessibility hurdles faced by members of minority populations in learning systems are well documented~\cite{hinsvark2021accented,tatman2017gender,koenecke2020racial,alsharhan2020investigating}. These hurdles can be attributed in part to lack of representation of minority groups in large datasets, but other factors also come into play, especially with smaller datasets and feature-engineered methods. For each trait, we split the data such that each split has the same amount of training and testing data.

\section{Experimental Results and Discussion}
To evaluate our trained models, we simulate object retrieval tasks with objects found in a sensed environment. The system is given a description and is responsible for selecting the correct objects from a subset of objects in \gld. In a real-world setting, descriptions will often match multiple objects in an environment; in the dataset, natural language utterances describe the image they are associated with but they are also likely to describe other images of the same object or different objects of the same class. One of the main advantages of our manifold alignment approach is the possibility of retrieving multiple visual embeddings that are within a threshold of a language embedding in the shared manifold. While the model should be able to select the target object given a description, it should also be able to separate negative and positive instances using this threshold. In order to evaluate the model against these two goals, we consider Mean Reciprocal Rank (MRR)-based and threshold-based evaluation tasks.

\setlength\dashlinedash{2pt}
\setlength\dashlinegap{2pt}
\setlength\arrayrulewidth{0.5pt}

\subsection{Learning Directly from Speech Improves Performance}
\label{ssec:mrr}
\begin{table}[ht]
\centering
\resizebox{\columnwidth}{!}{
\begin{tabular}{lcccc}
\hline
& Triplet MRR & Subset MRR & F1   \\ \hline
wav2vec 2.0 Speech     & \textbf{0.85 ($\pm$0.002)}       & \textbf{0.86 ($\pm$0.002)}     & \textbf{0.83 ($\pm$0.003)} \\
wav2vec 2.0 Trans.               & 0.83 ($\pm$0.002)       & 0.83 ($\pm$0.002)      & 0.79 ($\pm$0.003)\\
\hdashline
vq-wav2vec                    & 0.82 ($\pm$0.004)        & 0.78 ($\pm$0.004)      & 0.76 ($\pm$0.004) \\
DeCoAR                        & 0.80 ($\pm$0.003)        & 0.72 ($\pm$0.004)      & 0.71 ($\pm$0.003)\\
MFCC                          & 0.69 ($\pm$0.004)       & 0.49 ($\pm$0.01)             & 0.67 ($\pm$0)\\
\hdashline    Random Baseline & 0.61        & 0.46             & ---  \\ \hline
\end{tabular}
}
  \caption{MRR \& F1 Results (higher is better) with standard deviation over 5 runs. In addition to the queried object, the triplet setting includes an object from the same class and an object from a different class. The subset setting includes 4 objects from other classes. Our wav2vec 2.0 Speech approach achieves the strongest performance.
  \label{tab:mrr} \label{tab:f1}
  }
\end{table}
\subsubsection{Downstream Object Retrieval}
We simulate a robotic object retrieval task in which the goal is to retrieve the correct target vision instance $i$ for a given language utterance. The system has $N$ chances to pick an object given a description. The metric Mean Reciprocal Rank (MRR) measures how many tries are necessary for the correct object to be selected. We evaluate on the average of the reciprocal rank $\frac{1}{n_i}$ across all testing instances $\frac{1}{M}\sum_{i=1}^M \frac{1}{n_i}$. The reciprocal rank is the inverse of the rank at which the target object was retrieved.

The first retrieval setting is inspired by the triplets used for training. The MRR is calculated from a set of 3 objects: the target object, an object from the same class (but different instance) and an object from a different class. A real world example of this would be a robot picking between a green apple, a red apple and an orange when a description of a green apple is given. We will refer to the MRR performance from this setting as Triplet MRR. A limitation of this metric is that in cases of multiple positive examples (e.g., two green apples), the system is over-penalized. To counter the effects of this limitation, we consider a second setting that involves 5 objects: the target and 4 randomly selected objects from different classes. The MRR performance from this setting is referred to as Subset MRR.

We evaluate all 5 methods on these two object retrieval tasks and report the results in Table \ref{tab:mrr}. In these results, wave2vec 2.0 represents the state-of-the-art in speech featurizations, and outperforms other approaches.
As expected, learning a grounded language model from wav2vec 2.0 speech approach outperforms text transcriptions using the same featurizations. This confirms the hypothesis that the information lost in the transcription process negatively affects the performance of the grounding model. We note that adding another model for transcriptions to embeddings causes more latency between speech act and robot response. %
We find our baseline featurization, an MFCC, performs no better than chance; DeCoAR and the vq-wav2vec methods both achieve reliable results but are outperformed by wav2vec 2.0.

\subsubsection{Classification by threshold}

\begin{figure}
\centering
\adjustbox{max width=\columnwidth}{%
\input{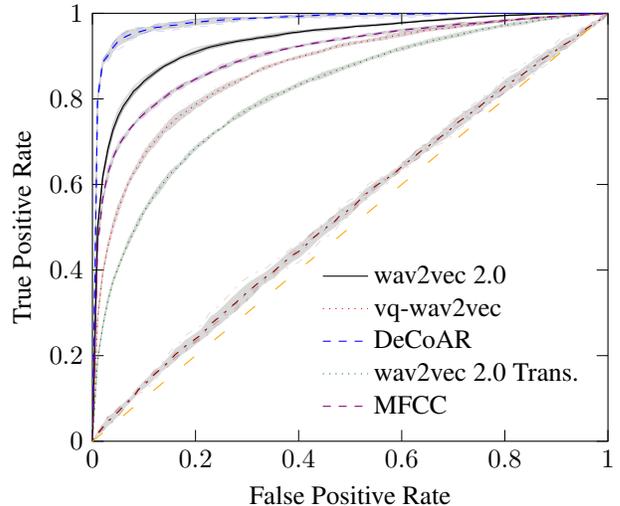}
}
\caption{We compare the ROC curves of each model on the validation set. The gray area around each curve represents the standard deviation of the model's performance over 5 runs. Higher AUC and closeness of the ROC curve to the top left corner mean that the model is better at discriminating between positive and negative examples of a given language description. The performance of the MFCC approach approximates that of a model with no skill.
}
\label{fig:roc_curves}
\end{figure}

\begin{figure}
\centering
\adjustbox{max width=\columnwidth}{%
\input{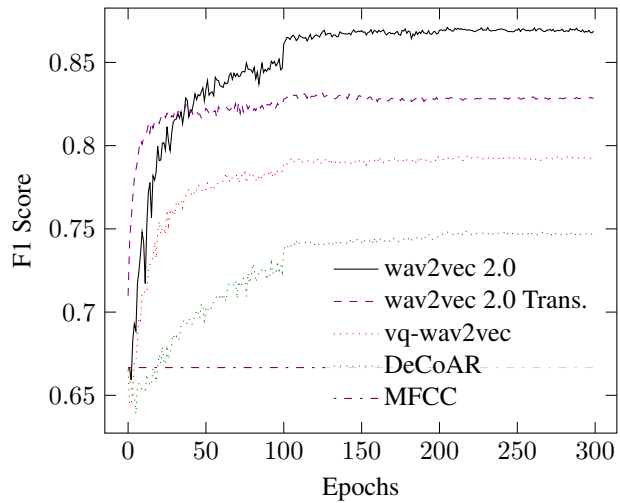}
}
\caption{We compare the convergence of F1 performance on the validation set as a function of training time (measured by training epochs). Each model's performance is averaged over 5 runs. Notice that the wav2vec 2.0 speech approach resulted in the best performance, followed by the transcriptions from wav2vec 2.0, while DeCoAR converges slightly faster than vq-wav2vec. The slight bumps at epoch 100 and 200 are due to a decrease in learning rate. MFCCs consistently underperformed.
}

\label{fig:f1_curves}
\end{figure}

The most intuitive way to deploy our alignment models is to define a fixed threshold $t$ such that any object within radius $t$ of a language utterance in the learned manifold is predicted to be described by that utterance. This can be defined as a binary classification task where an object falling within a radius $t$ of a language utterance is a positive prediction. We use our held-out data to simulate this task by considering every visual percept of the same class as a language description to be a positive instance and randomly sampling the same number of percepts from other classes to be negative instances. The F1 measure of this task is reported, as we value both the model's precision and its recall.

A key aspect of this problem is to determine the value of $t$. We tune the threshold $t$ for every model on the validation set. We divide the cosine distance by 2 to normalize the values between 0 and 1.
We find that a threshold of 0.4 achieves peak performance for the wav2vec 2.0 approaches. A threshold of 0.45 achieves peak performance for the vq-wav2vec and DeCoAR approaches. The MFCC approach achieves peak performance at a threshold of 1, which results in the model only making true predictions, since every instance---positive or negative---will be within a radius of 1 of the description. This indicates that the MFCC approach performs poorly regardless of threshold. The ROC curves in Fig. \ref{fig:roc_curves} confirm this assessment and show that all approaches except the MFCC approach learn to discriminate between positive and negative examples of language descriptions.

We apply the obtained thresholds to the testing set and report the results in \cref{tab:f1}. \Cref{fig:f1_curves} shows the evolution of the F1 scores of the 5 models on the validation set over the course of training.
Again, our wav2vec 2.0 Speech approach performed better than transcriptions, and the DeCoAR and vq-wav2vec methods achieve reasonable F1 scores of 0.71 and 0.76.
These results further support the conclusions reached in Section \ref{ssec:mrr}, confirming that grounded language acquisition from raw speech can lead to tangible results that surpass those of the traditional ``transcription-first'' method.

\subsection{Speaker Traits Study}

\subsubsection{Preprocessing}\label{sssec:singleuserlearning}

For each user, we train both a language alignment model and a vision alignment model with the manifold alignment learning objective described above. We exclude users from our dataset who did not provide at least 2 examples for at least 5 object classes. We further exclude the remaining users' examples from the object classes for which they have provided less than 2 examples. These constraints both guarantee that we include users who have provided sufficient examples for meaningful evaluation, and ensure there are 5 object classes for the subset MRR evaluation.
This leaves us with 87 users sub-selected from the complete dataset. These speakers provide an average count of 61.5 examples each, with a median of 35.

On average we trained on 40.7 examples. Even though we are taking advantage of domain encoding for speech, language, and vision, this small amount of training data per user is still a challenge.
On average we had 20.8 testing examples per user. We take this into account when analyzing the end performance of the model.
Of the 87 speakers, 50.5\% had accents. For gender, 39.1\% were annotated as men, 57.4\% as women, and 3.5\% as undetermined. 24.1\% of the users had creak, 4.6\% had hoarseness, 11.5\% had high levels of muffled-ness. 2.3\% of users had low volume, 82.8\% had medium volume, and 14.9\% had high volume. 90.8\% of users had low background noise and 9.2\% of users had high background noise.
We heard multiple kinds of background noise in the samples, including alarms, children, and fans. These noises contribute to real-world situation representation in the data. Due to the low amount of users with hoarse voices, we exclude hoarseness from the individual user study.

\subsubsection{Individual User-based Model Performance}

\begin{figure}
\vspace{2ex} %
\centering
\adjustbox{max width=\columnwidth}{%
\begin{tikzpicture}

\begin{axis}[
legend cell align={left},
legend style={fill opacity=0.8, draw opacity=1, text opacity=1, draw=none},
tick pos=both,
x grid style={white!69.0196078431373!black},
xmin=-0.79, xmax=7.79,
xtick style={color=black},
xtick={0,1,2,3,4,5,6},
xticklabel style = {rotate=30.0,anchor=base,yshift=-0.25cm,xshift=-0.75cm,font=\small},
xticklabels={No. Examples,Accent,Gender,Creak,Muffled-ness,Volume,Backg. Noise},
y grid style={white!69.0196078431373!black},
ymin=-0.75, ymax=1,
ytick style={color=black},
height=0.6\columnwidth,
width=\columnwidth,
]
\draw[draw=none,fill=black] (axis cs:-0.4,0) rectangle (axis cs:0,0.9079462724);
\addlegendimage{ybar,ybar legend,draw=none,fill=black}
\addlegendentry{Speech}

\draw[draw=none,fill=black] (axis cs:0.6,0) rectangle (axis cs:1,-0.3509467378);
\draw[draw=none,fill=black] (axis cs:1.6,0) rectangle (axis cs:2,0.004315509963);
\draw[draw=none,fill=black] (axis cs:2.6,0) rectangle (axis cs:3,-0.1598824835);
\draw[draw=none,fill=black] (axis cs:3.6,0) rectangle (axis cs:4,-0.005536944435);
\draw[draw=none,fill=black] (axis cs:4.6,0) rectangle (axis cs:5,0.3500293758);
\draw[draw=none,fill=black] (axis cs:5.6,0) rectangle (axis cs:6,-0.09523002141);
\draw[draw=none,fill=red] (axis cs:-2.77555756156289e-17,0) rectangle (axis cs:0.4,0.7678145415);
\addlegendimage{ybar,ybar legend,draw=none,fill=red}
\addlegendentry{Transcription}

\draw[draw=none,fill=red] (axis cs:1,0) rectangle (axis cs:1.4,-0.6073923676);
\draw[draw=none,fill=red] (axis cs:2,0) rectangle (axis cs:2.4,0.01345150136);
\draw[draw=none,fill=red] (axis cs:3,0) rectangle (axis cs:3.4,-0.04507484784);
\draw[draw=none,fill=red] (axis cs:4,0) rectangle (axis cs:4.4,-0.07035173242);
\draw[draw=none,fill=red] (axis cs:5,0) rectangle (axis cs:5.4,0.389522313);
\draw[draw=none,fill=red] (axis cs:6,0) rectangle (axis cs:6.4,-0.1388362432);
\end{axis}

\end{tikzpicture}
}
\caption{We compare the correlation between Subset MRR performance and different user qualities for the wav2vec2.0 speech and transcription methods. Accent is negatively correlated with performance in both, but the correlation is stronger when using transcriptions. The difference in performance is less pronounced for other speaker traits.
}
\label{fig:correlation_subset}
\end{figure}
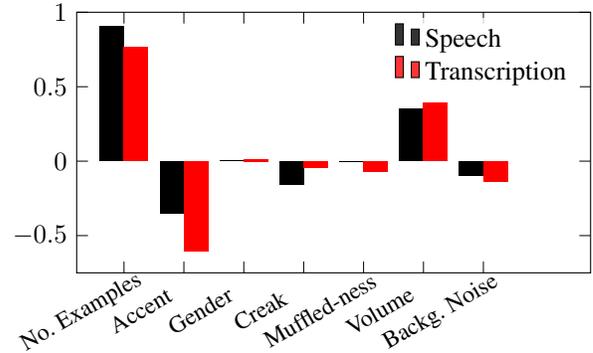

For each user, we train two models using the transcriptions and speech embeddings from the wav2vec 2.0 model. We look at the Pearson correlation coefficient (PCC) between each of the qualities we labeled and MRR scores. This analysis allows us to see which factors cause variation in both methods and which groups are most affected by the loss of information that occurs during transcription. The correlation results for the subset MRR are shown in Fig. \ref{fig:correlation_subset}. The results are mostly similar across both MRR metrics.

As expected, we find that for both methods, performance is negatively correlated with accent, creak and background noise and positively correlated with volume and the number of examples provided by the user. Background noise has a slight decreased correlation with the speech method compared to the correlation with the transcription method. This may be a benefit of not strictly mapping to a language token but rather a discretized high-dimensional value.

A key takeaway from the experiment is the significant gap between the correlation of the two models' performance with accented language, in which language models learned directly from accented speech are less negatively affected than those learned from transcriptions of that speech. In terms of the overall effects, the difference in subset MRR between the accented speakers and the non-accented ones for the transcription-first approach is triple the difference of the raw speech approach (approximately 6\% vs 2\%). This gap suggests that accented users are the most affected by the information loss of the transcription process. We expected that the more noisy nature of the speech representations provided by the raw speech method would help the learner in alleviating bias. The less negative correlation with accent supports this claim.

\subsubsection{Group-based Model Performance}

Finally, we split based on groups of users. We include workers who were excluded from the  individual-user study in these experiments, as aggregating the user data allows for proper training and test splits.
Data is split between accented and unaccented users, users with creak and without creak, perceived men and women, low (1), medium (2,3), and high volume (4), low (1,2) and high (3,4) background noise, and low (1,2) and high (3) muffled-ness. For each trait, we control for training and testing examples by splitting the data into equally sized groups with the same number of training and testing examples. As with the  individual-user study, we ensure that, for each group, the model is tested on object classes seen in training.

\begin{table}[t]

\begin{adjustbox}{max width=\columnwidth}
\begin{tabular}{lrrrrr}
            & \multicolumn{1}{c}{} & \multicolumn{2}{c}{Transcriptions} & \multicolumn{2}{c}{Speech}     \\\toprule
            & No. examples     & Subset   & Triplet   & Subset & Triplet \\\toprule
non-accent  & 8191                 & 0.82            & 0.85             & 0.84 & 0.85           \\
accent      & 8191                 & \textbf{0.92}   & \textbf{0.89}    & \textbf{0.93} & \textbf{0.90}  \\\midrule
non-creak  & 4932                 & 0.83            & 0.85             & 0.85 & 0.85           \\
creak      & 4932                 & \textbf{0.85}   & \textbf{0.86}    & \textbf{0.87} & \textbf{0.87}  \\\midrule
low volume  & 350                  & 0.62            & 0.75             & 0.55 & 0.73           \\
med. volume & 350                  & 0.64            & 0.77             & 0.57 & 0.72           \\
high volume & 350                  & \textbf{0.66}  & \textbf{0.78}    & \textbf{0.58} & \textbf{0.74}  \\\midrule
men         & 7897                 & 0.84   & 0.85    & 0.85 & 0.86  \\
women       & 7897                 & \textbf{0.89}& \textbf{0.87}             & \textbf{0.91} & \textbf{0.89}           \\\midrule
low backg.  & 1352                 & \textbf{0.74}   & \textbf{0.81}    & \textbf{0.68} & \textbf{0.79}  \\
high backg. & 1352                 & 0.73            & \textbf{0.81}             & \textbf{0.68} & \textbf{0.79}           \\\midrule
low muffl.     & 1610                 & \textbf{0.74}   & \textbf{0.81}    & \textbf{0.72} & \textbf{0.80}\\
high muffl.     & 1610                 & 0.73   & 0.80    & 0.70 & 0.79\\ \bottomrule
\end{tabular}
\end{adjustbox}
\caption{MRR scores for User Group-splits with wav2vec 2.0 transcriptions and speech methods. Higher is better. For both models, performance increases as volume goes from low to high at each tier. However, there is no decrease in performance for accented users.
 \label{tab:groups}
 }
\end{table}
We  trained and tested the wav2vec 2.0 speech and transcriptions methods on these splits. We report results in \cref{tab:groups}. For the volume splits, we saw a steady increase from low to high at each tier. For both models, the decrease in performance for accented users noticed in the individual-user study is absent. We capture the performance through statistics that utilize linear relationships to explain causes within our  individual-user study. We see that each participant has unique combinations of analyzed and non-analyzed characteristics play a factor on each axis but when grouping, the variance is addressed. We see this analysis as a critical step for further investigations into analysis and technical methods to support the study of bias within individual user understanding.

\section{Conclusion}

We have shown that it is both possible and effective to directly learn natural language groundings
from raw speech data to visual percepts, without having to rely on the intermediate textual representations of prior work. Our results demonstrate that direct grounding of speech to vision can minimize information loss and enable more reliable human-agent communication. Our investigation into direct grounding of speech includes a user study that identifies speaker/audio traits that historically affect speech recognition. We show that accented users are most affected by this information loss; direct grounding to raw speech has the potential benefit of reducing systems performance bias toward these and potentially other populations of users. While identifying and assigning these traits is preliminary, these initial results are relevant to bias and effectiveness in deployed, real-world systems.
In future work, we intend to resume demonstrating our results on a physical platform, when that once again becomes feasible.

\section*{Acknowledgments}
This material is based in part upon work supported by the National Science Foundation under Grant Nos. 1813223, 1920079, 1940931, and 2024878.

\appendix

\bibliography{main}
\clearpage

\end{document}